\documentclass[10pt,letterpaper]{article}
\usepackage[top=0.85in,left=2.75in,footskip=0.75in]{geometry}

\usepackage{authblk}

\usepackage{amsmath,amssymb}
\DeclareMathOperator*{\argmax}{arg{}max}
\DeclareOldFontCommand{\bf}{\normalfont\bfseries}{\mathbf}
\usepackage{changepage}

\usepackage{textcomp,marvosym}

\usepackage{cite}

\usepackage{nameref,hyperref}

\usepackage[right]{lineno}

\usepackage[nopatch=eqnum]{microtype}
\DisableLigatures[f]{encoding = *, family = * }

\usepackage[table]{xcolor}

\usepackage{array}

\newcolumntype{+}{!{\vrule width 2pt}}

\newlength\savedwidth

\raggedright
\usepackage[aboveskip=1pt,labelfont=bf,labelsep=period,justification=raggedright,singlelinecheck=off]{caption}

\makeatletter
\renewcommand{\@biblabel}[1]{\quad#1.}
\makeatother

\usepackage{lastpage,fancyhdr,graphicx}
\usepackage{epstopdf}
\fancyheadoffset[L]{2.25in}
\fancyfootoffset[L]{2.25in}
\begin{document}
\vspace*{0.2in}

\begin{flushleft}
{\Large
\textbf\newline{MMM and MMMSynth: Clustering of heterogeneous tabular data, and synthetic data generation} 
}
\newline
\\
Chandrani Kumari\textsuperscript{1,2},
Rahul Siddharthan\textsuperscript{1,2*},
\\
\bigskip

\textbf{1} The Institute of Mathematical Sciences, Chennai, India
\\
\textbf{2} Homi Bhabha National Institute, Mumbai, India
\\
\bigskip






* rsidd@imsc.res.in

\end{flushleft}
\section*{Abstract}
We provide new algorithms for two tasks relating to heterogeneous tabular datasets: clustering, and synthetic data generation. Tabular datasets typically consist of heterogeneous data types (numerical, ordinal, categorical) in columns, but may also have hidden cluster structure in their rows: for example, they may be drawn from heterogeneous (geographical, socioeconomic, methodological) sources, such that the outcome variable they describe (such as the presence of a disease) may depend not only on the other variables but on the cluster context. Moreover, sharing of biomedical data is often hindered by patient confidentiality laws, and there is current interest in algorithms to generate synthetic tabular data from real data, for example via deep learning.

We demonstrate a novel EM-based clustering algorithm, MMM (``Madras Mixture
Model''), that outperforms standard algorithms in determining clusters in synthetic heterogeneous data, and recovers structure in real data. Based on this, we demonstrate a synthetic tabular data generation algorithm, MMMsynth, that pre-clusters the input data, and generates cluster-wise synthetic data assuming cluster-specific data distributions for the input columns. We benchmark this algorithm by testing the performance of standard ML algorithms when they are trained on synthetic data and tested on real published datasets. Our synthetic data generation algorithm outperforms other literature tabular-data generators, and approaches the performance of training purely with real data.



\section*{Introduction}
Tabular datasets, consisting of heterogeneous variable types (categorical, ordinal, numeric), are ubiquitous in data science and, in particular, in biomedical research. Such datasets may additionally be heterogeneous in rows: they may consist of a mix of different subtypes or categories, corresponding to hidden structures not originally measured or are part of the dataset (such as geographical location, socioeconomic class, genotype, etc).

For machine learning applications, often one variable (often called a ``response'' or ``output''  variable) is of clinical interest (such as presence or absence of a disease or disorder such as diabetes, birth weight of a fetus) and the goal is to train a model to predict it from the other measurable variables (often called ``predictor '' or ``input'' variables). Patient confidentiality often restricts the ability to share such datasets freely, and several algorithms have been developed~\cite{xu2019modeling,li2020sync,xu2020synthesizing} to generate synthetic datasets that closely resemble real datasets and can be used to train ML models and shared freely. 

Here we address both tasks, of clustering heterogeneous data, and of generating realistic synthetic datasets as measured by their performance in training models on them for ML prediction on the real data.

\subsection*{Existing clustering methods}
Several standard clustering algorithms exist for multidimensional data and are implemented in machine learning packages such as \texttt{scikit-learn}~\cite{pedregosa2011scikit}, \texttt{ClusterR}\cite{clusterR} and \texttt{Clustering.jl}\cite{stukalov2021clustering}. These include agglomerative(hierarchical) clustering methods such as UPGMA, partitioning algorithms such as $K$-means that seek to optimize a particular distance metric across $K$ clusters, and mixture models such as Gaussian Mixture Models (GMMs) that assume that the data is drawn from a mixture of distributions and simultaneously learn the parameters of the distributions and the assignment of data points to distributions.

These methods assume an underlying appropriate distance metric (such
as Euclidean distance) (agglomerative clustering, or $K$-means), or
assume an underlying probability distribution for the data (GMMs)
which is to be learned.

Many datasets consist of a mixture of binary (eg, gender) categorical
(eg, ethnicity), ordinal (eg, number of children), and numerical data (eg, height, weight). Columns in such a tabular dataset may be correlated or interdependent. 

\subsection*{Overview of the MMM algorithm}

Here we propose an algorithm, which we call the Madras Mixture Model (MMM), to cluster tabular data where the columns
are assumed independently drawn from either categorical or real data. For each
clustering, we optimize the likelihood of the total data being drawn
from that clustering:
\begin{equation}
  P(D|K) = \prod_{j=1}^K P(d:d \in j)
\end{equation}
where $K$ is the number of clusters,  and the right hand side is a product over clusters for the likelihood that the subset of rows $d$ that are assigned to cluster $j$ would be co-clustered (this is stated more precisely in Methods). 

We assume that categorical variables are drawn from an unknown
categorical distribution with a Dirichlet prior, and numeric variables
are drawn from an unknown normal distribution with a normal-Gamma
prior. These are the conjugate priors for the
  multinomial and normal distributions respectively: that is, they are
  analytically tractable and the posteriors have the same functional
  form as the priors, while the form is flexible enough to represent a
  variety of prior distributions.

Unlike with standard GMM algorithms, we do not attempt to estimate the parameters
of the distributions, but integrate over them to obtain the
likelihoods (see Methods).
Our clustering approach is a variation of expectation maximization
(EM) where, essentially, the M-step in the usual GMM algorithm is
replaced with this integration.

To determine the true number of clusters, we use the
marginal likelihood (ML) (sometimes called the Bayesian Occam's razor)~\cite{mackay1992bayesian}. While the 
Bayesian information criterion\cite{neath2012bayesian}, is widely used as an approximation
to the ML, its performance on our synthetic dataset benchmarks was
inferior (see supplementary information). 

The most accurate numerical calculation of the ML is using thermodynamic integration (TI)~\cite{gelman1998simulating,lartillot2006computing}, reviewed in Methods, and we provide an implementation using TI. This is computationally expensive since it involves sampling at multiple different ``temperatures'' and integrating. As a faster alternative, 
the ML is frequently estimated using the harmonic mean (HM)
of samples\cite{newton1994approximate}, but this is known to give a biased estimate in
practice\cite{xie2011improving}. We give an improved approximation, which we call HM$\beta$, involving a fictitious inverse
temperature $\beta$, which, for suitable $\beta$, converges to the true value much faster
than the HM on small datasets where the exact answer is calculable,
and also converges rapidly to a fixed value on larger datasets. We
demonstrate that HM$\beta$ with $\beta\approx 0.5$ produces results comparable to TI on our synthetic datasets.

\subsection*{MMMSynth: generating synthetic tabular data}
  Patient confidentiality is both an ethical requirement in general
  and a legal requirement in many jurisdictions.  Clinical datasets
  may therefore only be shared if patients cannot be identified;
  ``pseudonymization'' (replacing real names with random IDs) is not
  enough in general since it may be possible to identify patients from
  other fields including age, clinical parameters, geographical
  information, etc.
  
Given the difficulty of sharing clinical datasets without violating
patient confidentiality, there has been interest in using machine
learning to generate synthetic data that mimics the charateristics of real
tabular data. Several methods have been
proposed~\cite{xu2019modeling,li2020sync,xu2020synthesizing}; these
generally rely on deep learning or other sophisticated approaches.

We use MMM as a basis for a relatively straightforward synthetic
tabular data generation algorithm, MMMsynth.  Each cluster is replaced
with a synthetic cluster which column-wise has the same statistical
properties for the input variable, and whose output variable is
estimated with a noisy linear function learned from the corresponding
cluster in the true data. All these synthetic clusters are then pooled
to generate a synthetic dataset.

We assess quality of synthetic data by the performance, in predicting
on real data, of ML models trained on synthetic data. We demonstrate
that this rather simple approach significantly outperforms other
published methods CT-GAN and CGAN, and performs, on average, better
than Gaussian Copula and TVAE. Our performance in many cases
approaches the quality of prediction from training on real data.

\section*{Materials and methods}

\subsection*{MMM: Clustering of heterogeneous data}

Consider a tabular data set consisting of $L$ heterogeneous columns and $N$ rows.
Each row of the set then consists of variables $x_i$, $i = 1, 2,
\ldots, L$. Each $x_i$ can be binary, categorical, ordinal, integer,
or real.  We consider only categorical (including binary) or numeric data; ordinal or numeric integer data can be treated as either categorical or numeric depending on context.  If there are missing data, they should first be
interpolated or imputed via a suitable method. Various
  imputation methods are available in the literature: for example,
  mean imputation, nearest-neighbour imputation, multivariate
  imputation by chained equations (MICE)~\cite{van1999flexible}.

\subsection*{Discrete data, Dirichlet prior}

For a categorical variable with $k$ values,
the Dirichlet prior is $P_(p) \propto \prod_{i=1}^k p_i^{c_i-1}$. For 
binary variables ($k=2$) this is called the beta prior. If we have already
observed data $D$ consisting of $N$ observations, with each outcome
$j$ occurring $N_j$ times, the posterior predictive for outcome $x=i$
($1 \le i, j \le k$) is

\begin{equation}
  P(x=i|D) = \frac{N_i + c_i}{N+C}
  \label{eq:postpred_categ}
\end{equation}

where $C = \sum_{i=1}^k c_i$.

\subsection*{Continuous data, normal-gamma prior}

For a continuous normally-distributed variable, we use a normal-gamma
prior, as described in~\cite{murphy2007conjugate},  with four hyperparameters, which we call $\mu_0, \beta_0, a_0,
b_0$:

\begin{align}
  p(\mu, \lambda) &= \mathcal{N}\left(\mu|\mu_0, (\beta_0\lambda)^{-1}
  \right) \mathrm{Gam}(\lambda|a_0,b_0)  \\
 &= \frac{(\beta_0\lambda)^{1/2}}{\sqrt{2\pi}}
  e^{-\frac{\beta_0\lambda}{2}(\mu-\mu_0)^2)}
  \frac{1}{\Gamma(a_0)}b_0^{a_0} \lambda^{a_0-1} e^{-b_0\lambda}
  \nonumber \\
 &= \left(\frac{\beta_0}{2\pi}\right)^{1/2}
  \frac{b_0^{a_0}}{\Gamma(a_0)} \lambda^{a_0-1/2}
  \exp\left(-\frac{\lambda}{2}\left[\beta_0(\mu-\mu_0)^2+2b_0\right]\right).
\end{align}

Here $\lambda$ is the inverse of the variance, $\lambda = \frac{1}{\sigma^2}$.  

Given data $D$ consisting of $n$ items $x_i$, $i = 1 \ldots n$, the posterior is
\begin{align}
  p(\mu, \lambda | D) &= NG(\mu, \lambda | \mu_n, \beta_n, a_n, b_n)
  \nonumber \\
& = \left(\frac{\beta_n}{2\pi}\right)^{1/2} 
\frac{b_n^{a_n}}{\Gamma(a_n)} \lambda^{a_n-1/2} \exp\left(-\frac{\lambda}{2}\left[\beta_n(\mu-\mu_n)^2+2b_n\right]\right)
\end{align}

where

\begin{align}
  \mu_n &= \frac{\beta_0 \mu_0 + n\bar{x}}{\beta_0 + n} \\
 \beta_n &= \beta_0 + n \\
 a_n &= a_0 + \frac{n}{2} \\
 b_n &= b_0 + \frac{1}{2} \sum_{i=1}^n (x_i - \bar{x})^2 
 + \frac{\beta_0 n(\bar{x}-\mu_0)^2}{2(\beta_0 +n)}
 \end{align}

The posterior predictive, for seeing a single new data item $x$ given
the previous data $D$, is~\cite{murphy2007conjugate}

\begin{equation}
  p(x|D) = \pi^{-1/2} \frac{\Gamma(a_n+\frac{1}{2})}{\Gamma(a_n)}\left(\frac{\Lambda}{2a_n}\right)^\frac{1}{2}
  \left(1+\frac{\Lambda(x-\mu_n)^2}{2a_n}\right)^{-\left(a_n+\frac{1}{2}\right)}
    \label{eq:postpred_normal}
\end{equation}

where 

\begin{equation}
  \Lambda = \frac{a_n \beta_n}{b_n(\beta_n+1)}
\end{equation}

In log space: 

\begin{equation}
  \begin{aligned}
     \log p(x|D) = -0.5 \log \pi + \log \Gamma(a_n+\frac{1}{2}) - \log \Gamma(a_n) + 0.5(\log \Lambda - \log (2a_n))\\
  -\left(a_n + \frac{1}{2}\right)
  \log\left(1+\frac{\Lambda(x-\mu_n)^2}{2a_n}\right)
  \end{aligned}
  \label{eq:postpred_norm}
\end{equation}

The marginal likelihood is
\begin{equation}
p(D) = \frac{\Gamma(a_n)}{\Gamma(a_0)} \frac{b_0^{a_0}}{b_n^{a_n}}
\left(\frac{\beta_0}{\beta_n}\right)^{1/2} (2\pi)^{-n/2}
\end{equation}

The posterior predictives for categorical (\ref{eq:postpred_categ})
  and normal (\ref{eq:postpred_norm}) are used in the clustering
  algorithm described in the next section.

\subsection*{Optimizing likelihood of a clustering by expectation maximization}

Let the data $D$ consist of $N$ rows, so that $D_i$ is the $i$'th
row. Let the model be denoted by $M_K$ where $K$ is the number of clusters.
Each cluster has its own parameters of the categorical or normal distribution
for each column which we call $\vec{\theta}$, with $\vec{\theta}_{\ell j}$ being the vector
of parameters for column $\ell$ in cluster $j$. The vector has $k-1$
independent components for a categorical distribution of $k$ categories, and $2$
components for a normal distribution, which are all
continuous. Another, discrete parameter for the model is the detailed
cluster assignment of each row $D_i$ to each cluster $C_j$. This can
be described by a vector $\vec{A}$ of length $N$, whose elements $A_i$
take values from $1$ to $K$.  A specific clustering is described by 
$\vec{\Theta} = \left\{  \vec{\theta}, \vec{A}\right\}$.

Given $K$, we seek an optimal clustering (i.e, optimal choice of
$\vec{A}$) by maximum likelihood. We write the likelihood being maximized is

\begin{align}
  \vec{A}^* &= \argmax_{\vec{A}} \prod_{j=1}^K P(d_j) \nonumber \\
  &= \argmax_{\vec{A}} \prod_{j=1}^K \int 
  P(d_j|\vec{\theta_j}) P(\vec{\theta_j})d\vec{\theta_j}. \label{eq:clustlik}
\end{align}


Here $d_j$ is shorthand for $\{D_i|A_i=j\}$, that is, it is the set
of rows from $D$ that belong to cluster $j$, and $\vec{\theta}_j$ is the set of parameters $\vec{\theta}$ specific to cluster $j$. The unknown parameters $\vec{\theta_j}$ are integrated over, separately for each cluster, and $P(\vec{\theta_j})$ is a Dirichlet or normal-Gamma prior as appropriate. That is, we seek to
find that assignment of individual rows to clusters, $\vec{A}$, such
that the product of the likelihoods that the set of rows $D_j$ that
have been assigned to the cluster $j$ are described by the same
probabilistic model is maximized. There is an implicit product over columns, which are assumed independent. 

In contrast to EM implementations of
GMMs, we only seek to learn $\vec{A}$; we do not seek the parameters
of the probabilistic model $\vec{\theta}$, but integrate over them.

An algorithm for clustering into $K$ clusters could be
\begin{enumerate}
\item Initialize with a random assignment of rows to clusters.
\item Calculate a score matrix $L_{ij}$. For each row $i$, this is the likelihood that $D_i$ belongs to
  cluster $j$ for all $j$ (excluding $i$ from its current
  cluster). 
\item Assign each row $i$ to the cluster corresponding to that value
  of $j$ which maximises $L_{ij}$. (This is similar to the E-step in EM.)
\item Repeat from step 2 until no reassignments are made.
\end{enumerate}

Instead of the M-step in EM, step 2 calculates likelihoods according to the posterior predictives for the categorical (\ref{eq:postpred_categ}) and normal (\ref{eq:postpred_normal}) distributions. The likelihood for a row is the product of the likelihoods over columns. We work with log likelihoods. 

In our implementation, we iterate starting from one cluster. After optimizing each $K$ clustering, starting at $K=1$, we use a heuristic to initialize $K+1$ clusters: we pick the poorest-fitting $\frac{N}{K+1}$ rows (measured by their posterior predictive for the cluster that they are currently in) and move them to a new cluster, and then run the algorithm as above.

We can choose to either stop at a pre-defined $K$, or identify the optimal $K$  via marginal likelihood, as described below (the optimal $K$ could be 1).

\subsection*{Identifying the correct $K$: Marginal likelihood}
Equation~\ref{eq:clustlik} maximizes the likelihood of a clustering over $\vec{A}$, while marginalizing over $\vec{\theta}$. The correct $K$ in the Bayesian approach is the $K$ that maximises the
 marginal likelihood (ML) marginalized over \emph{all} parameters
including $\vec{A}$. In other words, while one can increase the
likelihood in eq~\ref{eq:clustlik} by splitting into more and more clusters,
beyond a point
this will lead to overfitting; the full ML penalizes this (an approach also
called the Bayesian Occam razor~\cite{mackay1992bayesian}).  

Unfortunately,
exact calculation of the ML (marginalizing over $\vec{A}$)  is
impossible. There are several approaches using sampling; we review two
below (harmonic mean [HM], and thermodynamic integration [TI]), before
introducing our own, a variant of HM which we call HM$\beta$, which we
show is more accurate than HM, and on our data, comparably accurate to TI while being faster.

\subsubsection*{Arithmetic Mean and Harmonic Mean}

A straightforward estimation of the marginal likelihood for $K$
clusters ($ML_K$) would be to
sample uniformly from the parameter space for $\vec{A}$, and calculate the average likelihood
over $M$ samples (there is an implicit marginalization over $\vec{\theta}$ throughout):
\begin{equation}
  ML_K \equiv P(D|K) \approx \frac{1}{M} \sum_{m=1}^M P(D|K,\vec{A}_m)
\end{equation}
This is the arithmetic mean (AM) estimate, and tends to be biased to
lower likelihoods because the region of high-likelihood parameters is
very small.

An alternative is to start from Bayes' theorem:
\begin{equation}
  P(\vec{A}|D,K) = \frac{P(D|\vec{A},K) P(\vec{A}|K)}
  {\sum_{\vec{A'}}P(D|\vec{A'},K)P(\vec{A'}|k)}
\end{equation}
The denominator on the right is the marginal likelihood for $K$
clusters. Rearranging,
\begin{equation}
\frac{P(\vec{A}|K)}{ML_K} =
\frac{P(\vec{A}|D,K)}{P(D|\vec{A},K)}
\end{equation}
and summing over $\vec{A}$ with $\sum_{\vec{A}} P(\vec{A}|K) = 1$,
\begin{equation}
\frac{1}{ML_K} =
\sum_{\vec{A}} \frac{P(\vec{A}|D,K)}{P(D|\vec{A},K)} \label{eq:HM}
\end{equation}
If we sample $\vec{A}$ from the distribution $P(\vec{A}|D,K)$, then for $M$ samples we have
\begin{equation}
ML_K \approx \left( \frac{1}{M} \sum_{m=1}^M
\frac{1}{P(D|\vec{A}_m,K)} \right)^{-1}.
\end{equation}
This is the ``harmonic mean'' (HM) approximation. Both the HM and the
AM can be derived via different choices of an importance sampling
distribution in a Metropolis-Hastings
scheme~\cite{lartillot2006computing}. The HM is known to be biased towards
higher likelihoods in practice, oppositely to the AM.

\subsubsection*{Thermodynamic integration}
Thermodynamic integration (TI), a technique borrowed from physics, was
described in the statistical inference context by Gelman and
Meng \cite{gelman1998simulating}. The following quick summary is adapted to our
notation from Lartillot and Philippe \cite{lartillot2006computing}.

Suppose one has an un-normalized density in parameter space,
parametrized by $\beta$, $q_\beta(\vec{A})$. We can define a ``partition function''
\begin{equation}
  Z_\beta = \int q_\beta(\vec{A}) d\vec{A}. 
\end{equation}
In our case $\vec{A}$ is discrete, so the integral, here and below, should be interpreted as a sum. From this we get a normalized density 
\begin{equation}
  p_\beta(\vec{A}) = \frac{1}{Z_\beta} q_\beta(\vec{A}).
\end{equation}  
We then have
\begin{align*}
  \frac{\partial}{\partial \beta} \log Z_\beta &= \frac{1}{Z_\beta}
  \frac{\partial Z_\beta}{\partial \beta} \\
  &= \frac{1}{Z_\beta}  \frac{\partial}{\partial \beta} \int
  q_\beta(\vec{A}) d\vec{A} \\
  &= \int \frac{1}{q_\beta(\vec{A})} \frac{\partial
      q_\beta(\vec{A})}{\partial \beta} \frac{q_\beta(\vec{A})}{Z_\beta}
      dA \\
      &= E_\beta \left[ \frac{\partial \log q_\beta(\vec{A})}{\partial
          \beta} \right]
\end{align*}
Defining $U(\theta) = \frac{\partial}{\partial \beta} \log q_\beta(\vec{A})$, and integrating from $\beta = 0$ to $1$,
  \begin{equation}
    \log Z_1 - \log Z_0 = \int_0^1 E_\beta[U] d\beta.
  \end{equation}

Consider the particular choice
\begin{equation}
  q_\beta(\vec{A}) = P(D|\vec{A},K)^\beta P(\vec{A}|K)
\end{equation}
where, as above, $K$ is the number of clusters.
Then $q_0$ is the prior for $\vec{\theta}$, and
$q_1$ is proportional to the posterior. Therefore $Z_0$ is 1 (since
$q_0$ is normalized) and $Z_1$ is the marginal
likelihood. Substituting,
\begin{equation}
  \log \mathrm{ML} = \int_0^1 E_\beta\left[\log P(D|\vec{A},K)\right] d\beta
\end{equation}.

The expectation $E_\beta$ is calculated by sampling at various $\beta$
and the integral is found by Simpson's rule.

\subsubsection*{A faster approximation}

We return to the HM approximation (eq~\ref{eq:HM}). The problem is
that the distribution $P(\vec{A}|D,K)$ is strongly peaked around
the optimal parameters $\vec{A}$. To broaden the distribution we
can introduce a fictitious inverse temperature $\beta$ (not the same as in TI), and write
\begin{equation}
\frac{1}{ML_K} =
\sum_{\vec{A}} \frac{P(\vec{A}|D,K)^\beta P(\vec{A}|D,K)^{1-\beta}}{P(D|\vec{A},K)} \label{eq:HMb}
\end{equation}

But 
\begin{equation}
  P(\vec{A}|D,K) =
  \frac{P(D|\vec{A},K)P(\vec{A}|K)}{ML_K}
\end{equation}
and this gives 
\begin{equation}
\frac{1}{ML_K} = \sum_{\vec{A}} P(\vec{A}|D,K)^\beta
\left(\frac{P(\vec{A}|K)}{ML_K}\right)^{1-\beta}P(D|\vec{A},K)^{-\beta}
\end{equation}
and therefore
\begin{equation}
  ML_K^{-\beta} = \sum_{\vec{A}}
  P(\vec{A}|D,K)^\beta
  P(\vec{A}|K)^{1-\beta}P(D|\vec{A},K)^{-\beta}
\end{equation}
        
We can evaluate $ML_K$ from this by sampling from
\begin{equation}\tilde{P}(\vec{A}|D,K)
  \equiv P(\vec{A}|D,K)^\beta P(\vec{A}|K)^{1-\beta}
\end{equation}
(in
practice from $P(D|\vec{A},K)^\beta$ which is proportional to this) rather
than $P(\vec{A}|D,K)$. However, this distribution is not normalized. Let 
$\sum_{\vec{A}} \tilde{P}(\vec{A}|D,K) = \sum_{\vec{A}}
P(\vec{A}|D,K)^\beta P(\vec{A}|K)^{1-\beta} = Z \neq1$.
If we take $M$ samples from  $\tilde{P}(\vec{A}|D,K)$, then
\begin{align}
  ML_K^{-\beta} &= \frac{1}{M} \sum_{m=1 \atop \mathrm{samples~from~}\tilde{P}}^M
  P(D|\vec{A},K)^{-\beta} \times Z \\
  &\equiv \left< P(D|\vec{A},K)^{-\beta}\right>_{\tilde{P}} Z.
\end{align}

To estimate $Z$ we are forced to sample from the normalized
distribution $P(\vec{A}|D,K)$:
\begin{align}
  Z &= \sum_{\vec{A}} \tilde{P}(\vec{A}|D,K) = \sum_{\vec{A}} \frac{P(\vec{A}|D,K)^\beta
  P(\vec{A}|K)^{1-\beta}}{P(\vec{A}|D,K)}
  P(\vec{A}|D,K)\\
  &\equiv \left<  \frac{P(\vec{A}|D,K)^\beta
  P(\vec{A}|K)^{1-\beta}}{P(\vec{A}|D,K)}\right>_P\\
 &= \left< \frac{P(D|\vec{A},K)^{\beta-1} P(\vec{A}|K)^{\beta-1}}{ML_K^{\beta-1}}
  P(\vec{A}|K)^{1-\beta} \right>_P
\end{align}

Substituting,
$$ ML_K^{-\beta} = ML_K^{1-\beta} \left< 
P(D|\vec{A},K)^{-\beta}\right>_{\tilde{P}} \left< P(D|\vec{A},K)^{\beta-1}
\right>_P $$

$$ML_K^{-1} = \left<P(D|\vec{A},K)^{-\beta}\right>_{\tilde{P}}
\left<P(D|\vec{A},K)^{\beta-1}\right>_P$$

and finally
\begin{equation}
\log ML_K = - \log\left<P(D|\vec{A},K)^{-\beta}\right>_{\tilde{P}} -\log
\left<P(D|\vec{A},K)^{\beta-1}\right>_P
\end{equation}
where, again, $\left<\ldots\right>_{\tilde{P}}$ means an average over
$M$ samples from $P(D|\vec{A},K)^\beta$, and $\left<\ldots\right>_P$ means an
average over $M$ samples from $P(\vec{A}D,K)$. This expression reduces to the
arithmetic mean if $\beta=0$ and to the harmonic mean if $\beta=1$.
We refer to it as HM$\beta$.

\begin{figure}
    \centering
    \includegraphics[width=12cm]{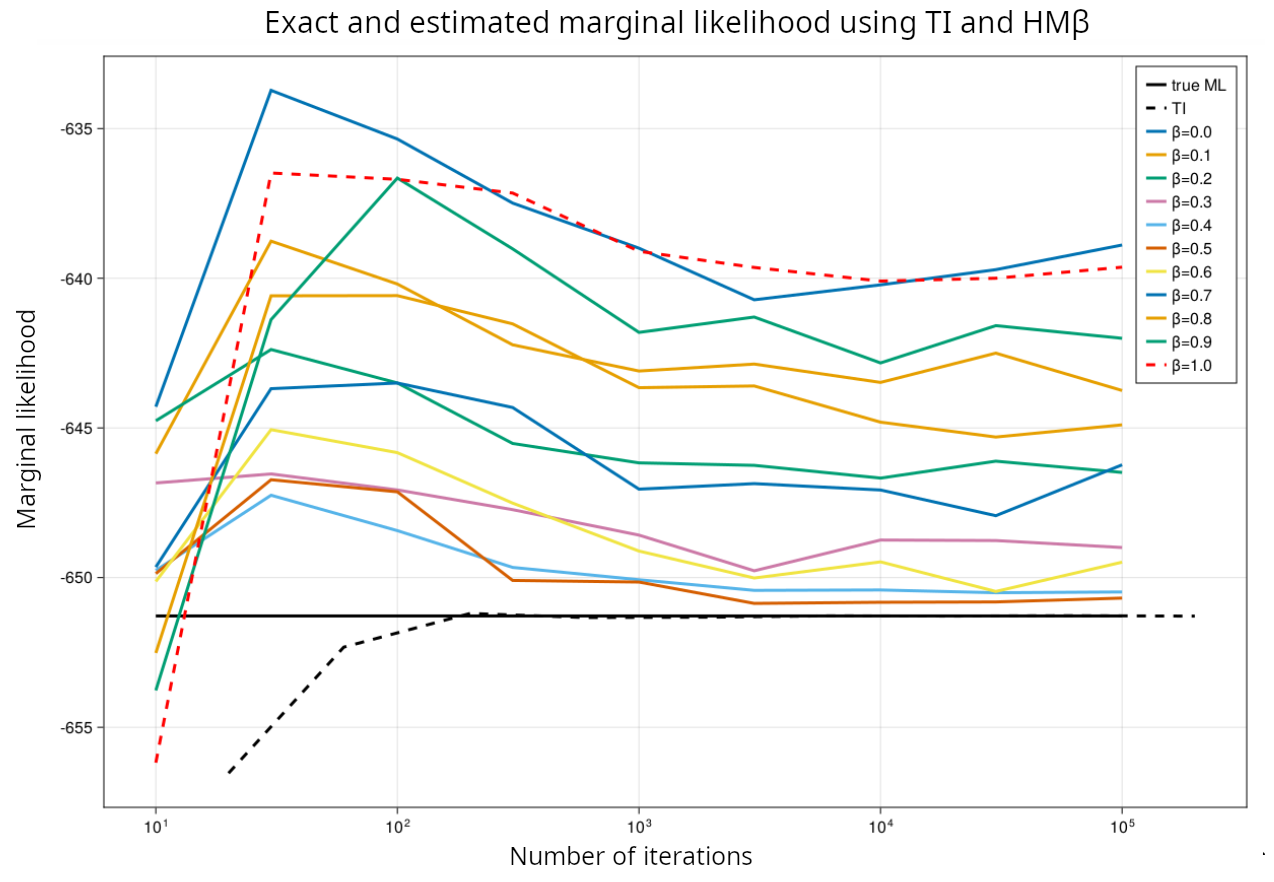}
    \caption{For a dataset of 20 rows, the marginal likelihood can be calculated exactly (solid line); this is compared with TI and with HM$\beta$ at various $\beta$. $\beta=0.5$ gives results close to the exact value.}
    \label{fig:exactml}
\end{figure}

We assess an optimal choice of $\beta$ using a small dataset of 20 rows where the marginalization can be carried out explicitly, by summing over all 1,048,574 possible cluster assignments to two clusters (figure~\ref{fig:exactml}). The TI method converges quite quickly to the exact answer, and the HM$\beta$ for $\beta=0.5$ also gives good results.  On larger datasets too we found $\beta=0.5$ an optimal choice (as in next subsection), though the exact value of the ML is not known. 

\subsubsection*{Synthetic data for benchmarking clustering}

We generated multiple datasets, each with 10 columns, 5000 rows, with
five clusters each, in a 5:4:3:2:1 ratio, with varying parameters, as
follows. For categorical data, columns in each cluster $j$ were
sampled from vectors $\boldsymbol{v}_0 + \Delta\boldsymbol{v}_j$ where
$\boldsymbol{v}_0$ was common to all clusters and $\boldsymbol{v}_j$
was specific to the $j$'th cluster, and $\Delta$ was varied from 0.5
to 4.5, with smaller values indicating greater similarity among
clusters. Here, five columns were binary and five were 4-valued. For
numeric data, the means in different clusters were separated by 1.0
and the standard deviations were separated by $\delta\sigma$, which
varied from 0.5 to 4.5, with smaller $\delta\sigma$ indicating
less dispersed, more distinct clusters (for small
  variance, clusters with different means have less overlap). We also
  simulated numeric data where the means in all clusters were the same
  and only the standard deviations differed by $\delta\sigma$. Finally, we generated mixed datasets, which included five numeric and five categorical (4-valued) columns, varying the parameter $\Delta$ as above and fixing $\delta\sigma = 5.0-\Delta$, so that increasing $\Delta$ implies increasing similarity among all columns. 

\subsection*{Assessment of benchmarking}
We used the adjusted Rand index  (ARI)~\cite{hubert1985comparing} to
compare true and predicted clusters.  

In supplementary information, we also show three other metrics:
Normalized Clustering Accuracy, Fowlkes-Mallows Score, and Adjusted
Mutual Information, as provided by the
GenieClust~\cite{gagolewski2021genieclust} suite; results were
similar. Some of these apply only when the true and predicted numbers
of clusters are the same. We also compare, for completeness, two
internal clustering metrics (which do not depend on a known
ground-truth clustering): silhouette score and Davies-Bouldin
score. These
require an internal distance metric to be defined for pairwise distances to be
defined; we use the Euclidean distance, but MMM does not use a metric,
but is based on a likelihood marginalized over hyperparameters, so
these measures are to be treated with caution.

\subsection*{MMMsynth: generating synthetic data with MMM}

\subsubsection*{Synthetic data generation algorithm}
MMMSynth uses MMM to pre-cluster an input dataset, excluding the output column. Each cluster is assumed, as in MMM, to consist of independent columns that are either categorical or numeric.  The parameters of the corresponding multinomial or Gaussian distribution are fitted to each column in each cluster, and a new cluster of the same size is generated by sampling from these distributions. A linear model is fitted to the output column in each real cluster and used to generate the output column in the synthetic clusters. The synthetic clusters are finally combined to produce a full dataset of the same size as the original dataset. 

For comparison we generated synthetic data using the methods available in literature. We used synthetic data vault\cite{SDV} libraries in python to generate synthetic data using TVAE, Gaussian Copula, CTGAN and CGAN.

\subsubsection*{Benchmarking MMMSynth-generated synthetic datasets}
To evaluate the similarity of the generated data of the real data, we trained machine learning models (logistic regression, random forest) on the synthetic data and evaluated their predictive performance on the real dataset. We also compared the performance of a model trained on the real dataset in predicting on the same dataset. We used six datasets from the UCI machine learning repository. 

\subsection*{Benchmarks: Real datasets used}

We used the following datasets from the UCI Machine Learning
Repository\cite{dua2017uci}, some of which were obtained via the
\href{https://www.kaggle.com}{Kaggle} platform:
\begin{itemize}
\item Abalone: predicting age of abalone from physical measurements, 8
  predictors (input variables), 1323 rows, from UCI
  \\ \href{https://archive.ics.uci.edu/ml/datasets/abalone/}{https://archive.ics.uci.edu/ml/datasets/abalone/}
  (note: original data had 4,177 rows and 28 output values, which are
  number of rings; for binary prediction we used only the rows with 9
  or 10 rings, which were the most frequent, resulting in a dataset of
  1323 rows of which 689 had 9 rings and 634 had 10 rings.)
\item Heart failure prediction dataset: compiled from UCI Machine
  learning Repository, 11 predictors, 918 rows, from
  \\ \href{https://www.kaggle.com/datasets/rishidamarla/heart-disease-prediction}{https://www.kaggle.com/datasets/rishidamarla/heart-disease-prediction}

\item  Pima Indians diabetes: 8 predictors, 768 rows, from \\ \href{https://www.kaggle.com/datasets/uciml/pima-indians-diabetes-database}{https://www.kaggle.com/datasets/uciml/pima-indians-diabetes-database}, source UCI
\item Breast cancer Wisconsin (Diagnostic) dataset: 30 predictors, 569 rows, from \\ \href{https://archive.ics.uci.edu/ml/datasets/Breast+Cancer+Wisconsin+%28Diagnostic%29}{https://archive.ics.uci.edu/ml/datasets/Breast+Cancer+Wisconsin+\%28Diagnostic\%29}
\item Maternal health risk data: 7 predictors, 676 rows, from
  \\ \href{https://www.kaggle.com/datasets/csafrit2/maternal-health-risk-data}{https://www.kaggle.com/datasets/csafrit2/maternal-health-risk-data}
       (note: original dataset had 1,014 rows with output
  values low, medium and high; for this task only 676 rows with output
  values low/high were retained.
\item Stroke prediction dataset: 10 predictors, 4909 rows, of which 209 positive and 4700 negative, from \href{https://www.kaggle.com/datasets/zzettrkalpakbal/full-filled-brain-stroke-dataset}{https://www.kaggle.com/datasets/zzettrkalpakbal/full-filled-brain-stroke-dataset}
\item Connectionist Bench (Sonar, Mines vs. Rocks) dataset (60 predictors, 207 rows, from \\\href{https://archive.ics.uci.edu/dataset/151/connectionist+bench+sonar+mines+vs+rocks}{https://archive.ics.uci.edu/dataset/151/connectionist+bench+sonar+mines+vs+rocks}
\end{itemize}

All of these have binary output variables. For clustering benchmarking, all were used (sonar was used as part of the ClustBench set, below). For synthetic data, the stroke dataset was omitted since it was highly unbalanced.

In addition, datasets from the UCI subdirectory of ClustBench~\cite{gagolewski2022framework} were used, many of which have multi-valued outputs. These consist of sonar (described above), as well as
\begin{itemize}
    \item ecoli: 7 predictors, 335 rows, 8 categories
    \item glass: 9 predictiors, 213 rows, 6 categories
    \item ionosphere: 34 predictors, 350 rows, 2 categories
    \item statlog: 19 predictors, 2309 rows, 7 categories
    \item wdbc: 30 predictors, 568 rows, 2 categories
    \item wine: 13 predictors, 177 rows, 3 categories
    \item yeast: 8 predictors, 1483 rows, 10 categories
\end{itemize}

\section*{Results}
\subsection*{Clustering algorithm performance}

\subsubsection*{Synthetic data: purely categorical, purely normal,
  mixed}

We generate three kinds of synthetic datasets: purely categorical,
purely numeric (normally distributed), and mixed, as described in
Methods. Each dataset has five clusters. We vary a parameter
($\delta\sigma$ for numeric, $D$ for categorical), as described in
Methods, to tune how similar the clusters are to each other. Large
$\delta\sigma$ indicates a larger variance, and greater overlap among
the ``true'' clusters; likewise small $D$ indicates that the
categorical distributions of different clusters are closer. However,
for numeric data, if clusters have the same mean, we expect that
larger $\delta\sigma$ would improve the ability to separate
them.

\begin{figure}
    \centering
    \includegraphics[width=13cm]{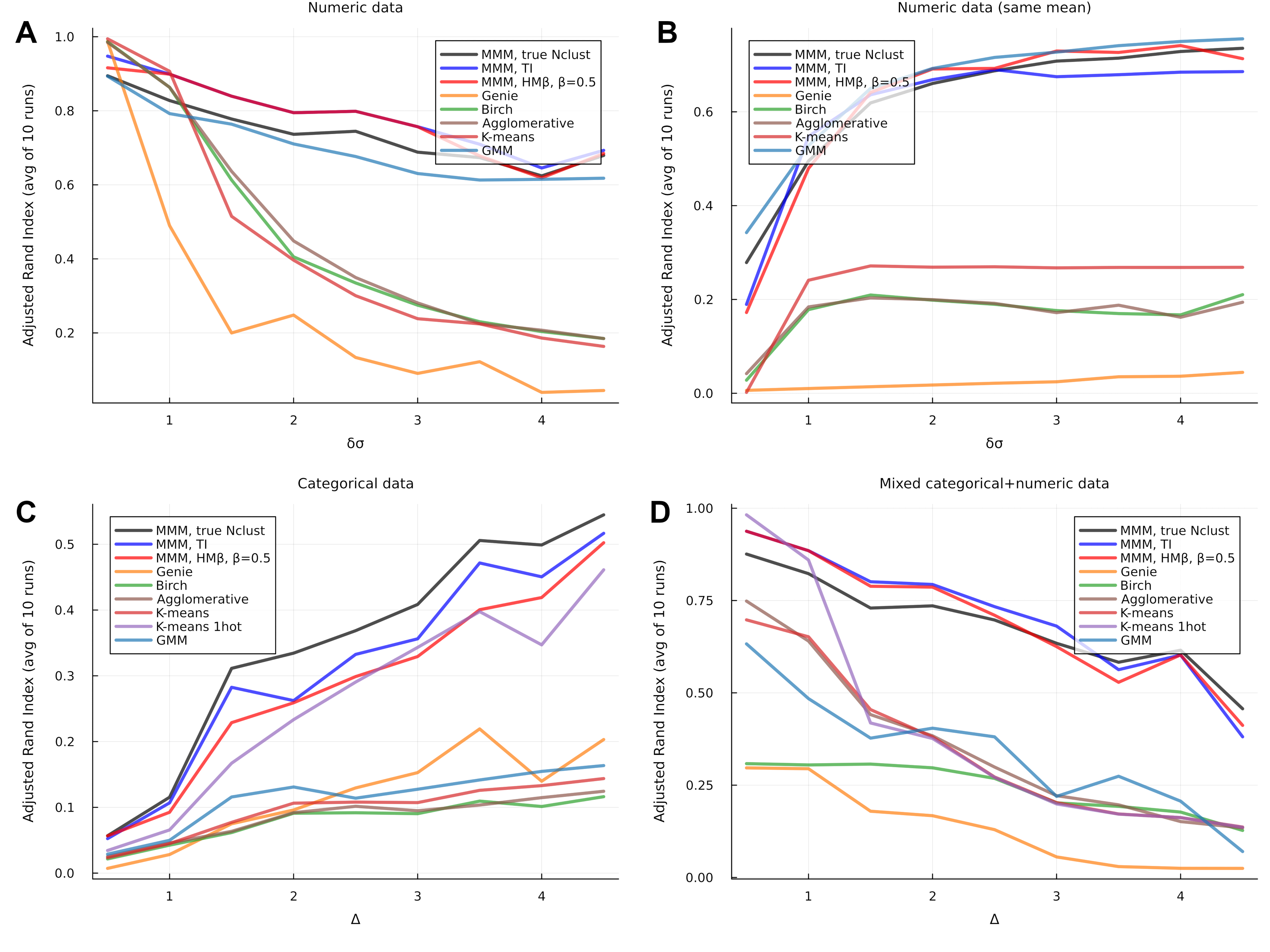}
    \caption{Clustering of four kinds of synthetic datasets: (A) purely numeric (normally distributed, differing means and variances), (B) purely numeric (normally distributed, same mean but differing variances), (C) purely categorical, and (D) mixed.}
    \label{fig:synthdata}
\end{figure}

Figure~\ref{fig:synthdata} shows the results for normalized accuracy:
for purely normally distributed data MMM performs comparably with
Gaussian mixture models, for purely categorical data we greatly
outperform all methods, with only $K$-means with one-hot encoding
coming close, and for mixed normal+categorical data, too, our
performance is superior to other methods. We run MMM in three modes:
telling it the true cluster size, or using the TI and HM$\beta$
methods (with $\beta=0.5$). All other methods are told the correct
cluster size.

  For numeric data with differing means, and for
  categorical data, MMM outperforms
  all other methods, but GMM comes close. For numeric data with the
  same mean and differing variances, GMM slightly outperforms MMM. For
  categorical data, K-means with 1-hot encoding comes close in
  performance to MMM. In dealing with mixed categorical+numeric data,
  MMM clearly outperforms all methods. 

  It is worth noting that the slopes of the ARI curves are opposite in
  figures \ref{fig:synthdata} A/D (differing mean) and B (uniform mean).
  This is because, when the means are different, a small variance
  makes the clusters more distinct and a large variance increases the
  overlap, but when the means are the same, the clusters can only be
  distinguished by their variance, so a larger $\delta\sigma$ helps.

  MMM with the true Nclust does not always appear to perform best by
  this metric, but the performance with true Nclust, TI and HM$\beta$
  are comparable in all cases.

\subsubsection*{Predicting the true number of clusters}
\begin{figure}
    \centering
    \includegraphics[width=13cm]{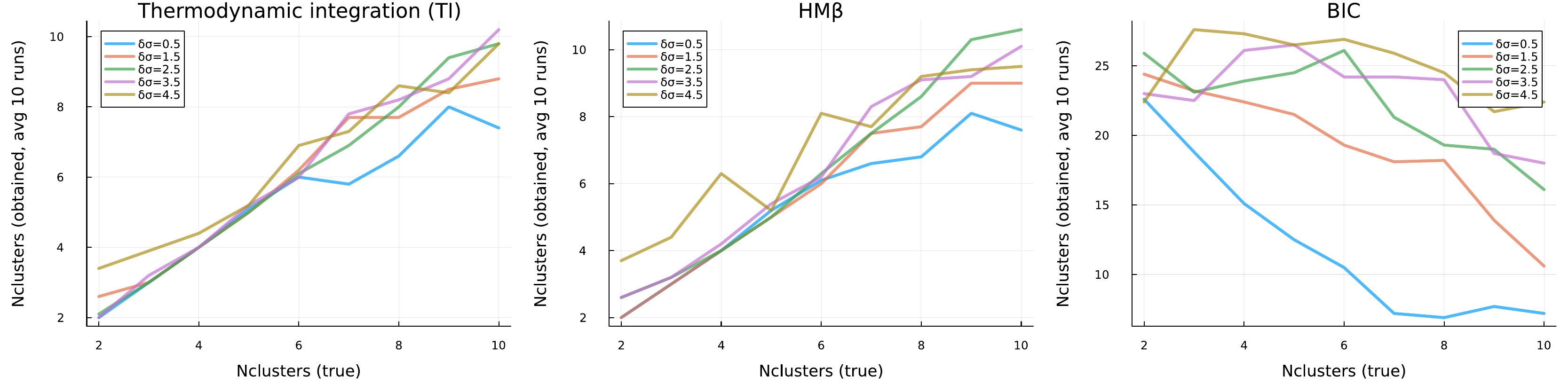}
    \caption{Optimal number of clusters obtained by TI, HM$\beta$ and
      Bayesian Information Criterion (BIC), on mixed
      categorical+numeric data with true cluster number ranging from 2
      to 10. TI and HM$\beta$ show
      comparably good results while, in our data, BIC is mostly unable
      to predict the true number of clusters.}
    \label{fig:nclust}
\end{figure}

We generated mixed categorical + numeric data, similar to above with
5000 rows per file, but divided into equal-sized clusters of 2 to 10
clusters. FIgure~\ref{fig:nclust} shows the performance of TI and
HM$\beta$ in predicting the true number of clusters; both perform
comparably, while the Bayesian information criterion (BIC) performs
poorly here, likely accounting for its poor performance in the
previous benchmark (not shown).

\subsubsection*{Real data}

\begin{figure}
\centering
\includegraphics[width=13cm]{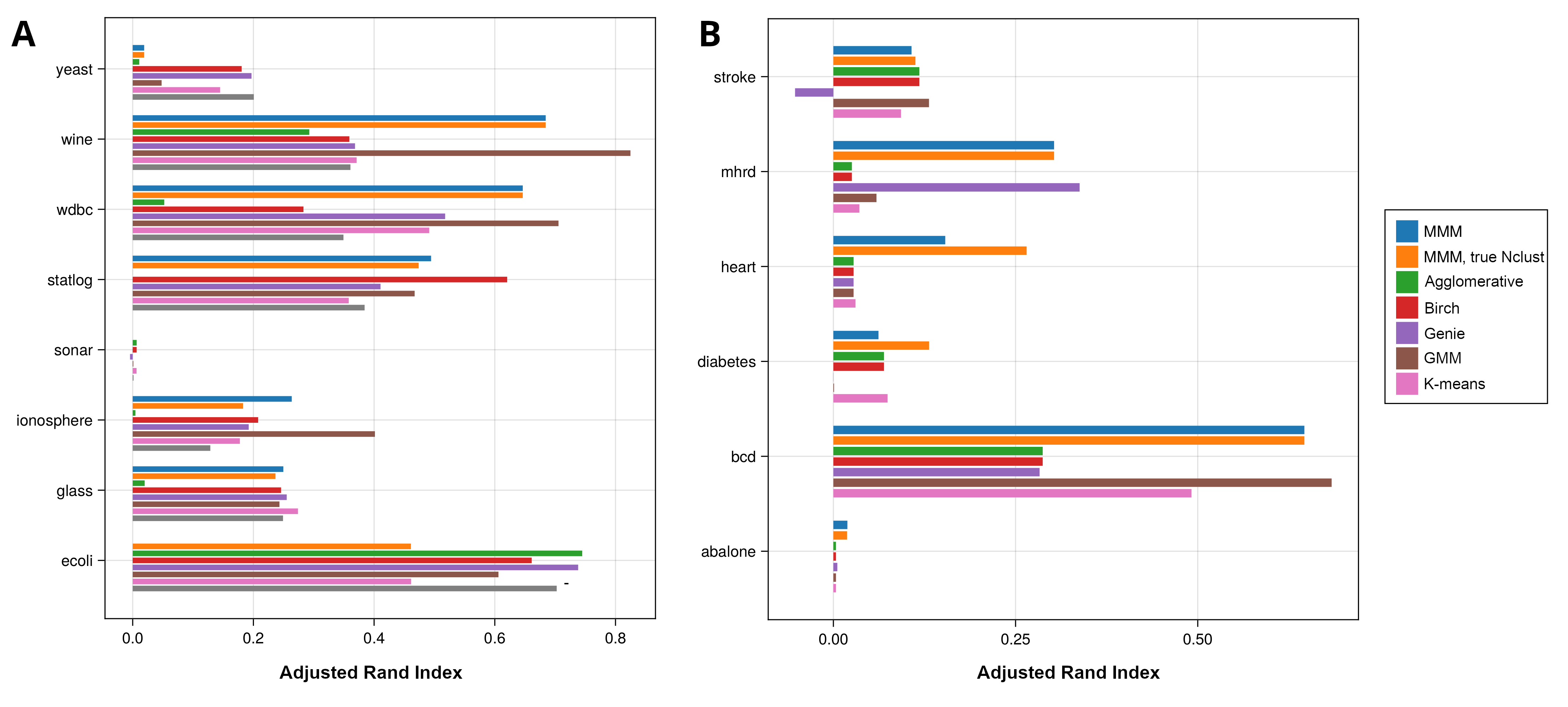}
\caption{Performance of MMM, using the TI criterion, and ``MMM, true Nclust'' where it is told the true number of clusters versus other programs, on eight datasets from ClustBench (A) and six datasets from UCI (B). Performance is reported using Adjusted Rand Index.}
        \label{fig:clustbench_kaggle}
\end{figure}

We consider the UCI abalone, breast, diabetes, heart, MHRD and stroke datasets described in Methods, which have binary outcome variables, and also eight datasets from the UCI machine learning database which are included in ClustBench~\cite{gagolewski2022framework}, which have output labels of varying number from 2 to 10. These are intended as tests of classification, not clustering, tasks. 
Nevertheless, clustering these datasets on input variables (excluding
the outcome variable) often shows significant overlap with the
clustering according to the output variable, as shown in
figure~\ref{fig:clustbench_kaggle}. This suggests that we are
recovering real underlying structure in the data. We compare various
other methods, all of which were run with the correct known number of
clusters, while MMM was run with (``MMM, true Nclust'') and without
(``MMM'' telling it the true number of clusters.

  The MMM results are with TI; the results with HM$\beta$ are similar and
  not shown for space reason. 

  Compared to the synthetic data results, several methods outperform
  MMM (even with fixed Nclust) on individual datasets, and MMM
  performs poorly on yeast, while all methods perform poorly on a few
  others. However, in most cases, MMM is among the better performers;
  in some cases the best performers, and in other cases inferior to
  another method, but no method is consistently superior. 
  
  When not told the true number of clusters, in some cases MMM refuses
  to cluster the dataset at all (as per the TI calculation of marginal
  likelihood, a single cluster is more likely than two). This is the
  case with the ecoli dataset. 

\subsubsection*{MMMSynth: ML performance on synthetic data}

\begin{figure}[!ht]
    \centering
    \includegraphics[scale=0.35]{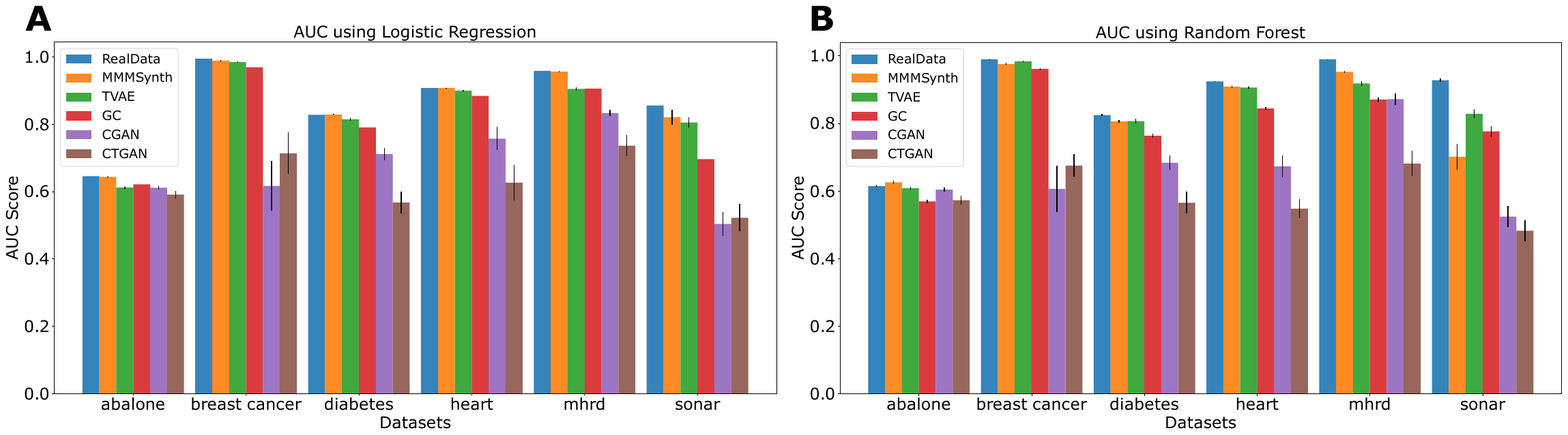}
    \caption{Logistic regression (A) and random forest (B) models were
      trained on the real data and on synthetic data generated using
      MMMSynth, TVAE, GC, CGAN and CTGAN and their predictive
      performance evaluated on the real datasets. The AUC (area under
      ROC curve, averaged over 20 runs) is shown for each method and
      each dataset, and errorbars are shown too.}
    \label{fig:auc}
\end{figure}

We use the six kaggle/UCI datasets (Abalone, Breast Cancer, Diabetes,
Heart, MHRD and Sonar) benchmarked above to generate synthetic data,
as described in Methods, train ML algorithms (logistic regression,
random forest) on these, and measure the predictive performance on the
real data. Figures ~\ref{fig:auc} show the results. Also tested are
Triplet-based Variational Autoencoder (TVAE), Gaussian Copula (GC),
Conditional Generative Adversarial Network (CTGAN), Copula Generative
Adversarial Network (CGAN).
AUC are averaged over 20 runs in each case.
In all datasets, we show performance
comparable to training on real data. TVAE and GC also perform well on
most datasets, while CGAN and CTGAN show poorer performance. All
programs were run with default parameters.

\section*{Discussion}

We present a clustering algorithm, MMM, that clusters heterogeneous data consisting of categorical and numeric columns. We demonstrate good performance on a variety of publicly available datasets and on our synthetic data benchmarks.  Speed optimizations will be explored in future work.  

Currently the columns are assumed to be independent, but it will be a
straightforward exercise to use a multivariate Gaussian to describe
the numeric columns. This too will be explored in future. Despite
this, our performance is comparable to and sometimes better than
scikit-learn's GMM implementation on the real-world UCI and clustbench
data presented here. In some benchmarks, other methods (Birch for
statlog, Agglomerative and Genie for ecoli) clearly outperform both us
and GMM. We emphasise again that the labels in these benchmarks are
not ground truth clustering labels, but output labels in
classification datasets. While (as commonly recognized), there is no
universal ``best'' clustering method, we perform well across a wide
variety of synthetic and real-data benchmarks, and further
improvements are possible. in the future.

We further use MMM as a basis for a synthetic data generation
algorithm, MMMSynth. We demonstrate that MMMSynth generates synthetic
data that closely resembles real datasets. Our method performs better
than current synthetic data generation algorithms in the literature
(TVAE, GC, CGAN and CTGAN), though the gap with TVAE in particular is
small.  Notably, all these methods explicitly
assume and model correlations between input columns: CGAN and GC use
copula functions to capture correlations between variables, while all
methods other than GC (CGAN, TVAE and CTGAN) employ deep learning
using all columns. In contrast, we first cluster the data and then
assume that, within each cluster, columns are uncorrelated. Our method
requires modest computational resources and no deep learning, and we
expect it will improve with improvements to the underlying MMM
clustering algorithm.

Our approach indirectly accounts for some correlations: for example, if two binary columns are correlated (1 tends to occur with 1, and 0 with 0), we would be likely to cluster the 1's together and 0's together. It will also account for multimodal numeric columns since these would be better represented as a sum of Gaussians. In tests on synthetically-generated non-normally-distributed numeric data (for example, Gamma-distributed with long tails), MMM breaks the data into multiple clusters, suggesting an attempt to approximate the Gamma distribution as a sum of Gaussians. This will be explored in a future work. Nevertheless, we do not see such a proliferation of clusters when running on real datasets. 

Currently, missing data needs to be imputed via another
  algorithm such as nearest-neighbour or MICE imputation. MMMSynth
  could be the basis for an alternative imputation algorithm in the
  future.

\section*{Acknowledgements}
This work grew from a project on personalized clinical predictions, for which we acknowledge discussion and collaboration with Gautam Menon, Uma Ram, Ponnusamy Saravanan, and particularly Leelavati Narlikar with whom we extensively discussed this work and whose insights were invaluable. We also acknowledge useful discussions with Durga Parkhi on synthetic data generation. 

\section*{Funding}
We acknowledge funding from BIRAC grant bt/ki-data0404/06/18 (RS), and the IMSc Centre for Disease Modelling (ICDM) funded via an apex project at IMSc by the Department of Atomic Energy, Government of India (CK, RS). The funders had no role in the data collection, research, analysis, writing or submission of the manuscript. 

\section*{Author contributions}
\begin{flushleft}

{\bf Conceptualization:} CK, RS \\
{\bf Data Curation:} CK, RS\\
{\bf Formal Analysis:} CK, RS\\
{\bf Funding Acquisition:} RS\\
{\bf Investigation:} CK, RS\\
{\bf Methodology:} CK, RS\\
{\bf Project Administration:} RS\\
{\bf Resources:} RS\\
{\bf Software:} CK, RS\\
{\bf Supervision:} RS\\
{\bf Validation:} CK, RS\\
{\bf Visualization:} CK, RS\\
{\bf Writing -- Original Draft Preparation:} CK, RS\\
{\bf Writing -- Review \& Editing:} CK, RS\\
\end{flushleft}

\section*{Code availability}
MMM and MMMsynth are available on \href{https://github.com/rsidd120/MadrasMixtureModel}{https://github.com/rsidd120/MadrasMixtureModel} under the MIT licence. They are implemented in Julia. 

\section*{Declaration of interests}
The authors declare no competing interests.

\nolinenumbers

%
%
%





\bibliography{references}

\end{document}